\documentclass[
10pt,
a4paper,
oneside,
headinclude,footinclude,
BCOR5mm,
]{scrartcl}
\RequirePackage{silence}
\usepackage[
nochapters, 
beramono, 
eulermath,
pdfspacing, 
dottedtoc 
]{classicthesis} 

\usepackage{arsclassica} 

\usepackage[T1]{fontenc} 

\usepackage[utf8]{inputenc} 

\usepackage{graphicx} 
\graphicspath{{Figures/}} 

\usepackage{paralist}

\usepackage{enumitem} 

\usepackage{lipsum} 

\usepackage{subfig} 

\usepackage{amsmath,amssymb,amsthm} 

\usepackage{varioref} 

\theoremstyle{definition} 

\theoremstyle{plain} 

\theoremstyle{remark} 

\hypersetup{
colorlinks=true, breaklinks=true, bookmarks=true,bookmarksnumbered,
urlcolor=webbrown, linkcolor=RoyalBlue, citecolor=webgreen, 
pdftitle={}, 
pdfauthor={\textcopyright}, 
pdfsubject={}, 
pdfkeywords={}, 
pdfcreator={pdfLaTeX}, 
pdfproducer={LaTeX with hyperref and ClassicThesis} 
}

\usepackage{hyphenat}

\usepackage{csquotes}
\PassOptionsToPackage{%
    backend=biber,%
    language=auto,%
    style=alphabetic,%
    maxcitenames=2,%
    backref=true,%
    urldate=edtf,%
    seconds=true,%
    date=edtf,%
    datamodel=custom,%
    datecirca=true,%
    dateuncertain=true,%
}{biblatex}
\usepackage{biblatex}
\DeclareBibliographyAlias{software}{online}
\DeclareLanguageMapping{english}{american-apa}
\DeclareSourcemap{
    \maps[datatype=bibtex]{
        \map[overwrite=true]{
            \step[fieldsource=author, match=Wienke, final]
            \step[fieldset=keywords, fieldvalue=ownpub]
        }
        \map{
            \step[fieldsource=organization, final]
            \step[fieldset=usera, origfieldval]
        }
        \map{
            \pertype{standard}
            \step[typesource=standard, typetarget=manual]
            \step[fieldset=keywords, fieldvalue={,type:standard}, append]
        }
        \map{
            \step[fieldset=abstract, null]
        }
        \map{
            \pernottype{software}
            \pernottype{online}
            \pernottype{report}
            \pernottype{techreport}
            \pernottype{standard}
            \pernottype{manual}
            \pernottype{misc}
            \step[fieldset=url, null]
            \step[fieldset=urldate, null]
        }
        \map{
            \pertype{inproceedings}
            \step[fieldset=venue, null]
            \step[fieldset=eventdate, null]
            \step[fieldset=eventtitle, null]
            \step[fieldset=isbn, null]
            \step[fieldset=volume, null]
        }
        \map{
            \pertype{software}
            \step[fieldset=author, null]
        }
    }
}
\newrobustcmd*{\citefullauthor}{\AtNextCite{\DeclareNameAlias{labelname}{first-last}}\citeauthor}
\DeclareLabelalphaTemplate{
    \labelelement{
        \field[final]{shorthand}
        \field{label}
        \field[strwidth=3,strside=left,ifnames=1]{labelname}
        \field[strwidth=1,strside=left]{labelname}
        \field[strwidth=3]{usera}
    }
    \labelelement{
        \field[strwidth=2,strside=right]{year}
    }
}
\newcommand{\citesoftware}[1]{\citetitle{#1}~\autocite{#1}}

\DeclareTextFontCommand{\texthtt}{\ttfamily\hyphenchar\font=45\relax}

\usepackage{tabularx} 
\newcommand{\tableheadline}[1]{\multicolumn{1}{c}{\spacedlowsmallcaps{#1}}}
\usepackage{caption}

\usepackage{longtable}
\usepackage{soul}
\usepackage{hyphenat}

\usepackage[binary-units=true]{siunitx}
\usepackage[abbreviations,british]{foreign}

\bibliography{report}

\title{\normalfont\spacedallcaps{{\normalsize Results of the survey}\protect\\Failures in Robotics and Intelligent Systems}}

\author{\spacedlowsmallcaps{Johannes Wienke* \& Sebastian Wrede*}}

\date{\small{}Version: 1.0, \today}

\begin{document}


\renewcommand{\sectionmark}[1]{\markright{\spacedlowsmallcaps{#1}}} 
\lehead{\mbox{\llap{\small\thepage\kern1em\color{halfgray} \vline}\color{halfgray}\hspace{0.5em}\rightmark\hfil}} 

\pagestyle{scrheadings} 


\maketitle 


\section*{Abstract} 

In January 2015 we distributed an online survey about failures in robotics and intelligent systems across robotics researchers.
The aim of this survey was to find out which types of failures currently exist, what their origins are, and how systems are monitored and debugged -- with a special focus on performance bugs.
This report summarizes the findings of the survey.

\setcounter{tocdepth}{1} 

\tableofcontents 

\listoffigures 

\listoftables 


{\let\thefootnote\relax\footnotetext{* \textit{Research Institute for Cognition and Robotics (CoR-Lab) \& Center of Excellence Cognitive Interaction Technology (CITEC), Bielefeld University, Germany. Contact: }\texthtt{\{jwienke,swrede\}@techfak.uni-bielefeld.de}}}

\newpage 

\section{Introduction}

Despite strong requirements on dependability in actual application scenarios, robotics systems are still known to be error prone with regular failures.
However, not many publications exist that have systematically analyzed this situation.
Therefore, we have decided to carry out a survey to get an assessment of the current situation in research robotics\footnote{Parts of the results have previously appeared in \textcite{Wienke2016}}.
The aim of this survey was to collect the impressions of robotics developers on the reliability of systems, reasons for failure, and tools used to ensure successful operation and for debugging in case of failures.
The survey specifically focused on software issues and software engineering aspects.
Apart from general bugs, performance bugs have been specifically addressed to understand their impact on robotics systems and to determine how performance bugs differ from other bugs.
A considerable amount of work in this direction has been done in other computing domains like high\hyp{}performance computing or for cloud services~\autocite[\eg{}][]{Gunawi2014,Jin2012,Zaman2012}.
However, in robotics such work is missing.
To our knowledge, only \textcite{Steinbauer2013} presents a systematic study on general faults in robotics systems, but without a specific focus on performance aspects.

Our survey was implemented as an online questionnaire (following methodology advices from~\textcite{GonzalezBanales2007}) which was distributed around robotics researchers using the well\hyp{}known mailing lists euRobotics (\texthtt{euron\hyp{}dist})\footnote{\url{https://lists.iais.fraunhofer.de/sympa/info/euron-dist}} and \texthtt{robotics\hyp{}worldwide}\footnote{\url{http://duerer.usc.edu/mailman/listinfo.cgi/robotics-worldwide}} as well as more focused mailing lists.
The detailed structure of the survey can be found in \autoref{app:survey}.
Please refer to this appendix for details on the phrasing of questions and permitted answers.
Results presented in the following sections are linked to the respective questions of the survey.

\begin{figure}[ht]
    \centering
    \includegraphics[width=0.9\linewidth]{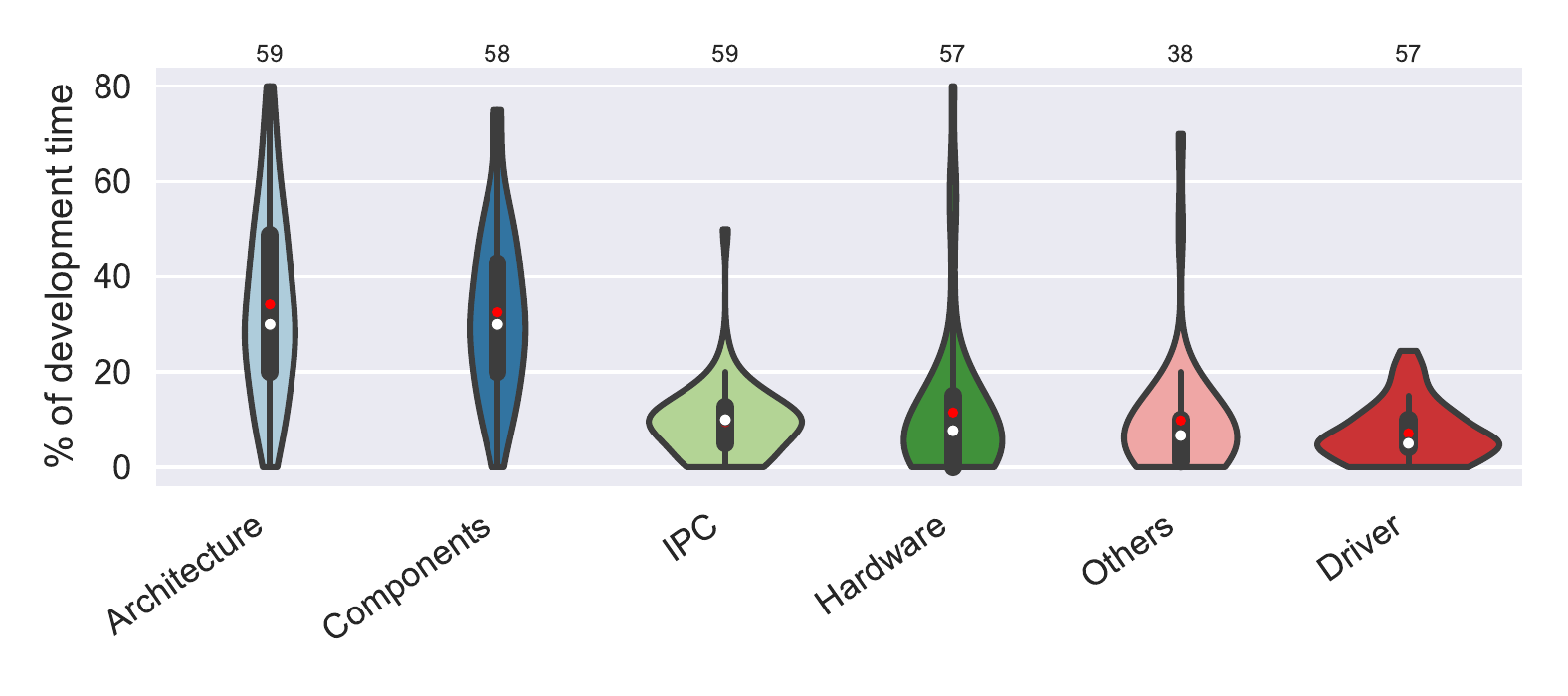}
    \caption[Participant development time spent on different aspects]{Development time spent by survey participants on different development aspects of robotics systems.
        Individual answers have been normalized to sum up to \SI{100}{\percent}.
        Inside the violins, a box plot is shown with the white dot representing the median and the red dot the mean value.
        Numbers above the plot express the sample size, which differs as answers were optional.}
    \label{fig:survey-domains}
\end{figure}

In total, \num{61} complete submissions and \num{141} incomplete ones\footnote{Incomplete submissions also include visitors who only opened the welcome page and then left.} were collected.
\SI{86}{\percent} of the participants were researchers or PhD candidates at universities, \SI{7}{\percent} regular students and \SI{7}{\percent} from an industrial context (\ref{questionnaire:section-94-question-context}).
On average, participants had \num{5.8} years of experience in robotics (sd: \num{3.3}, \ref{questionnaire:section-94-question-experience}).
Participants spend their active development time primarily with software architecture and integration as well as component development, despite individual differences visible in the broad range of answers (\cf{} \autoref{fig:survey-domains}, \ref{questionnaire:section-94-question-perc}).
Other activities like hardware or driver development are pursued only for a limited amount of time.

\section{Tool usage}
\label{sec:tool_usage}

\begin{figure}[t]
    \centering
    \includegraphics[width=0.9\linewidth]{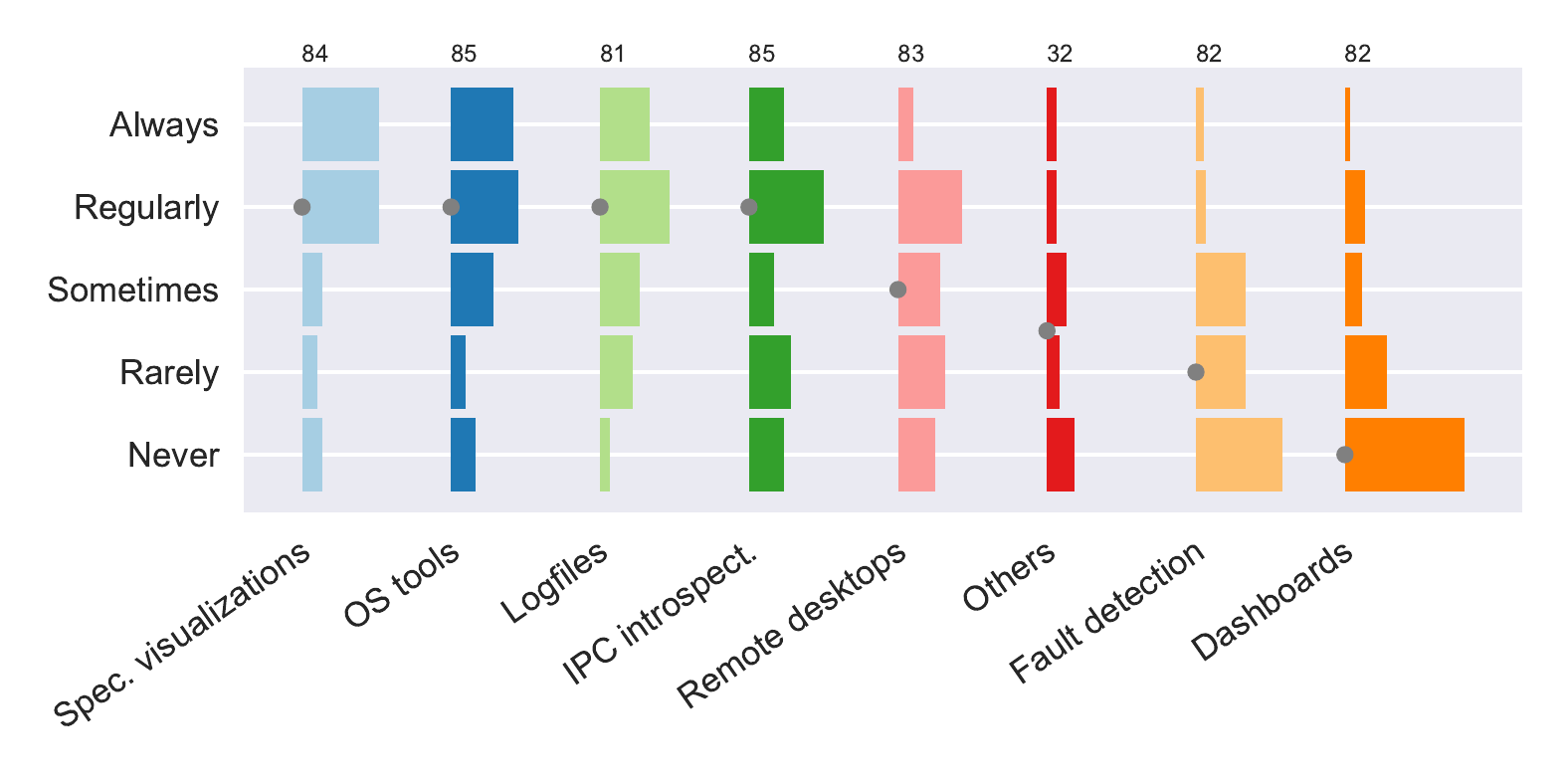}
    \caption[Monitoring tools usage frequencies]{Usage frequency for different categories of monitoring tools.
        For each category, the answer counts are displayed as a histogram and the grey point marks the median value.
        The categories are ordered by median and -- if equal -- mean values.
        Numbers above the plot express the sample size, which might differ as answers were optional.}
    \label{fig:survey-monitoring-tools}
\end{figure}

A first set of questions tried to assess which software tools are used to monitor and debug robotics systems in general.
For different types of tools, participants could rate on a 5 point scale from 0 (Never) to 4 (Always), how often the respective type of tool is used during usual development and operation of their systems.
For general monitoring tools (\ref{questionnaire:section-88-question-monitoringkinds}) the answers are depicted in \autoref{fig:survey-monitoring-tools}.
According to the developers' opinion, special purpose visualization tools like \citesoftware{rviz} or debug windows for image processing operations are most frequently used to monitor systems.
These are followed by low\hyp{}level operating system level tools like \texttt{ps} or \texttt{htop} and logfiles.
Tools related to distributed systems like utilities of the IPC mechanism form the last category of tools that is regularly used.
Remote desktop connections are used only sometimes.
In contrast, autonomous fault detection methods and special dashboards for visualizing system metrics are only rarely used, despite the fact that such tools are well\hyp{}established for operating large\hyp{}scale systems with high dependability requirements.

A second question regarding monitoring tools asked for the exact names of tools that are used (\ref{questionnaire:section-88-question-monitoringtools}).
The answers to this question are summarized in \autoref{app:used_monitoring_tools}.
The most frequently mentioned category of tools matched the previous question (visualization tools, most notably \citesoftware{rviz}).
These tools are followed by middleware\hyp{}related tools, most notably the ROS command line tools and \texthtt{rqt}, as well as operating system tools with \texttt{htop} and \texttt{ps} being the most frequently mentioned examples.
Finally, manual log reading, remote access tools, custom mission\hyp{}specific tools, and generic network monitoring tools like \citesoftware{wireshark} are used.
Additionally, one participant also explicitly mentioned hardware indicators like LEDs for this purpose.

\begin{figure}[t]
    \centering
    \includegraphics[width=0.9\linewidth]{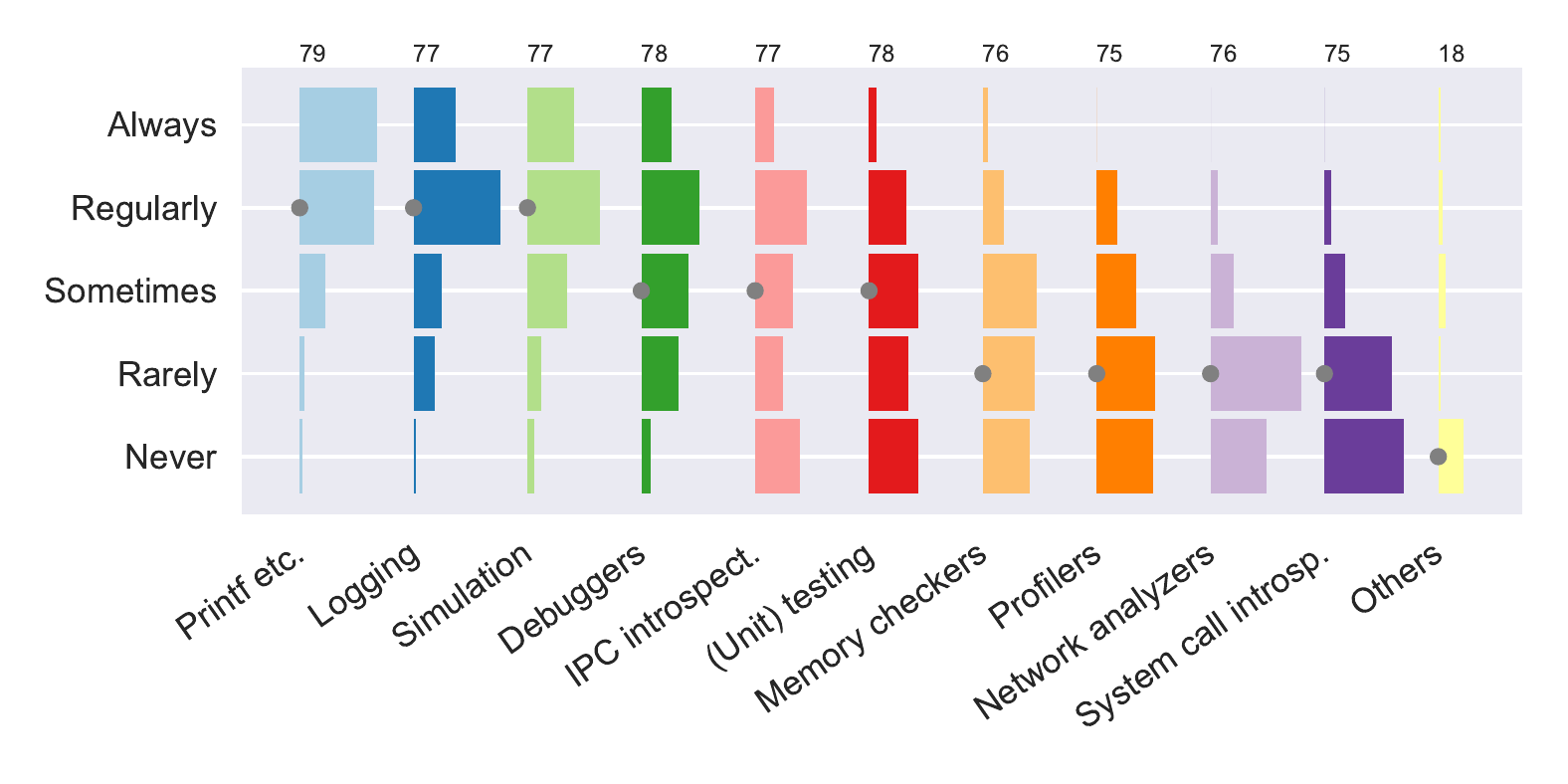}
    \caption[Debugging tools usage frequencies]{Usage frequencies for different categories of debugging tools and methods.}
    \label{fig:survey-debugging-tools}
\end{figure}

Regarding tools used to debug robotics systems (\ref{questionnaire:section-85-question-debuggingkinds}), participants mostly use basic methods like \texttt{printf} or log files as well as simulation (\cf \autoref{fig:survey-debugging-tools}).
General\hyp{}purpose and readily available debuggers are less frequently used than these basic methods.
Unit testing seems to be partially practiced and accepted in the robotics and intelligent systems community.

The actual tools being used have been summarized in \autoref{app:used_debugging_tools} as a result of question~\ref{questionnaire:section-85-question-debuggingtools}.
Debuggers represent the most frequently mentioned category of tools with \citesoftware{gdb} leading this category.
Another frequently used debugging tool is \citesoftware{valgrind} for checking memory accesses.
Besides \texttt{printf} debugging, other categories of used tools comprise middleware utilities, simulation and visualization (with gazebo being the most frequently mentioned software), and unit testing.

\section{Bugs and their origins}
\label{sec:bugs_and_origins}

\begin{figure}[t]
    \centering
    \includegraphics[width=0.6\linewidth]{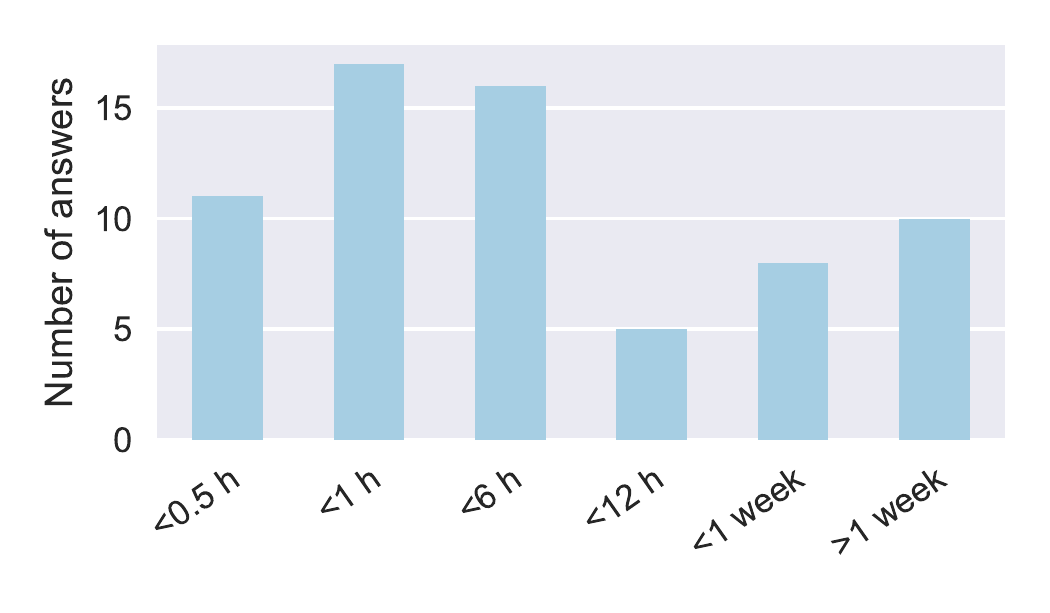}
    \caption[Observed MTBF]{Participant answers for the observed MTBF in their systems.}
    \label{fig:survey-mtbf}
\end{figure}

In a second set of questions we have addressed the reasons for and effects of bugs in robotics systems.
As actual numbers for failure rates in robotics systems are rarely available, one question asked participants for the MTBF they have observed in systems they are working with (\ref{questionnaire:section-86-question-mtbf}).
As visible in \autoref{fig:survey-mtbf}, the answers form a bimodal distribution where one part of the participants rates MTBF of their systems to be within the range of minutes to a few hours, whereas others indicate MTBF rates in the range of days to weeks.
One can think of multiple explanations for these diverging replies:
\begin{itemize}
    \item The systems participants have been working with are different in nature and some are closer to production systems.
    \item Answers with higher MTBF include the system's idle time in the calculation, despite an explicit indication in the explanation of the question that the \emph{operation} time is the basis for this number.
    \item Differences can be explained by the way people use debugging or monitoring tools in their systems.
        However, no significant relations could be found in the results.
\end{itemize}
As for the first two hypotheses no data is available to validate them and the third one cannot be proven using the survey results, the effective reasons for the bimodal distribution are unknown.

To generally understand why systems fail, participants were asked to rate how often different bug categories were the root cause of system failures (\ref{questionnaire:section-86-question-bugorigin}).
The categories have been selected based on related survey work from robotics and other domains~\autocite{Steinbauer2013,Gunawi2014,Jin2012,McConnell2004}.
\begin{figure}[t]
    \centering
    \includegraphics[width=0.9\linewidth]{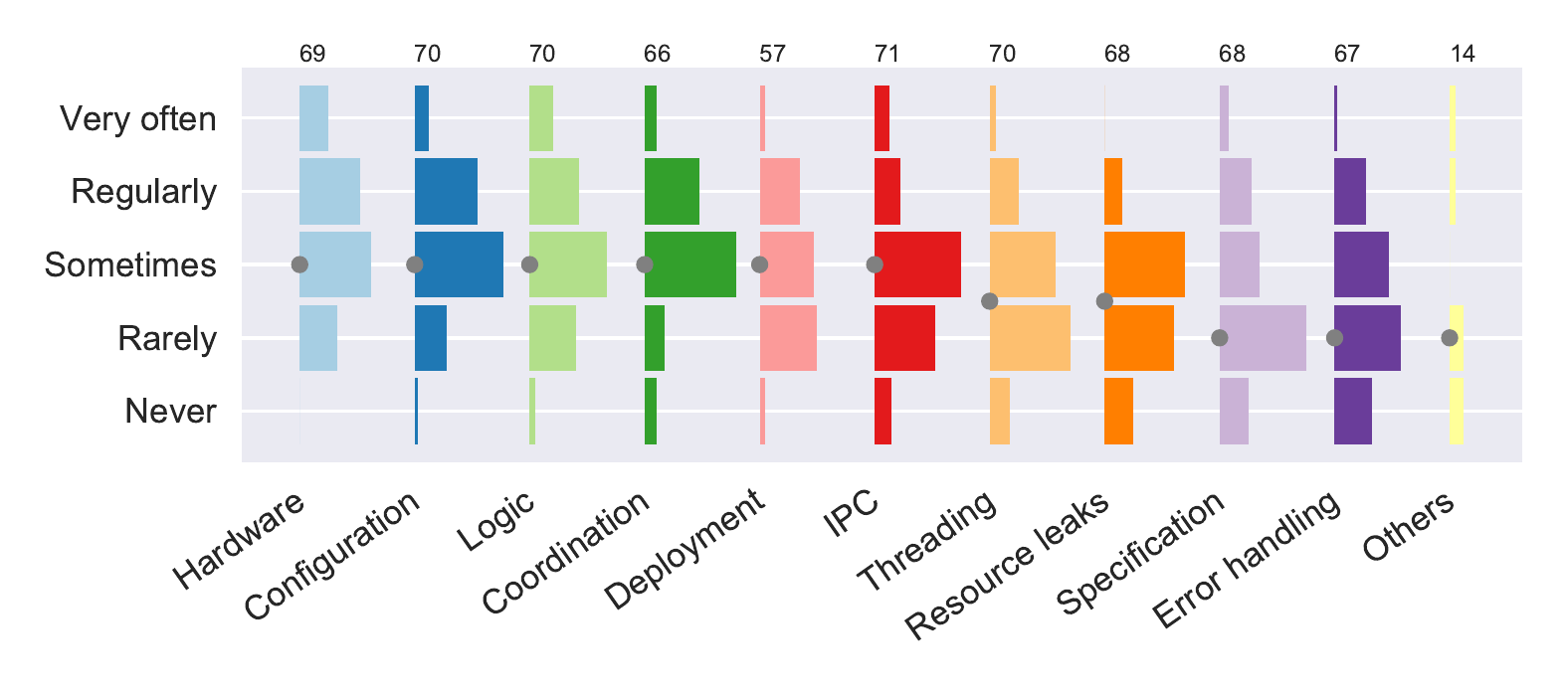}
    \caption[Frequencies of system failure reasons]{Rating of different bug categories being the reason for system failures.}
    \label{fig:survey-bug-origins}
\end{figure}
\autoref{fig:survey-bug-origins} displays the results for this question.
Hardware bugs represent the most frequent category followed by a set of categories representing high\hyp{}level issues (configuration, coordination, deployment, IPC) as well as general logic bugs.
Most of the high\hyp{}level issues seem to be technical problems and not specification problems because specification issues only rarely cause failure (median).

Apart from the aforementioned categories, participants could provide further causes in text form (\ref{questionnaire:section-86-question-bugoriginother}).
After removing items that relate to categories already presented in the previous question, answers can be summarized as
\begin{inparaenum}[a)] 
    \item environment complexity\fshyp{}changes (8 mentions)
    \item low\hyp{}level driver and operating system failures (3 mentions)
    \item hardware configuration management (1 mention) and
    \item hardware limitations (1 mention).
\end{inparaenum}
\autoref{app:summarization_of_free_form_bug_origins} shows the answers in detail as well as how categories have been assigned.
In the survey, we explicitly excluded the environment as an origin of system failures because it does not represent a real defect in any component of the system.
However, the results still show how important the discrepancy between intended usage scenarios and capabilities of systems in their real application areas is in robotics and intelligent systems.

\section{Performance bugs}
\label{sec:performance_bugs}

In order to understand performance bugs in robotics and intelligent systems, a dedicated set of questions was added to the survey.
First, participants were asked for the percentage of bugs that affected
resource utilization (\ref{questionnaire:section-87-question-perfbugrate}).
On average, \SI{24}{\percent} (sd \SI{17}{\percent}) of all bugs affected system resources.
\begin{figure}[t]
    \centering
    \includegraphics[width=0.9\linewidth]{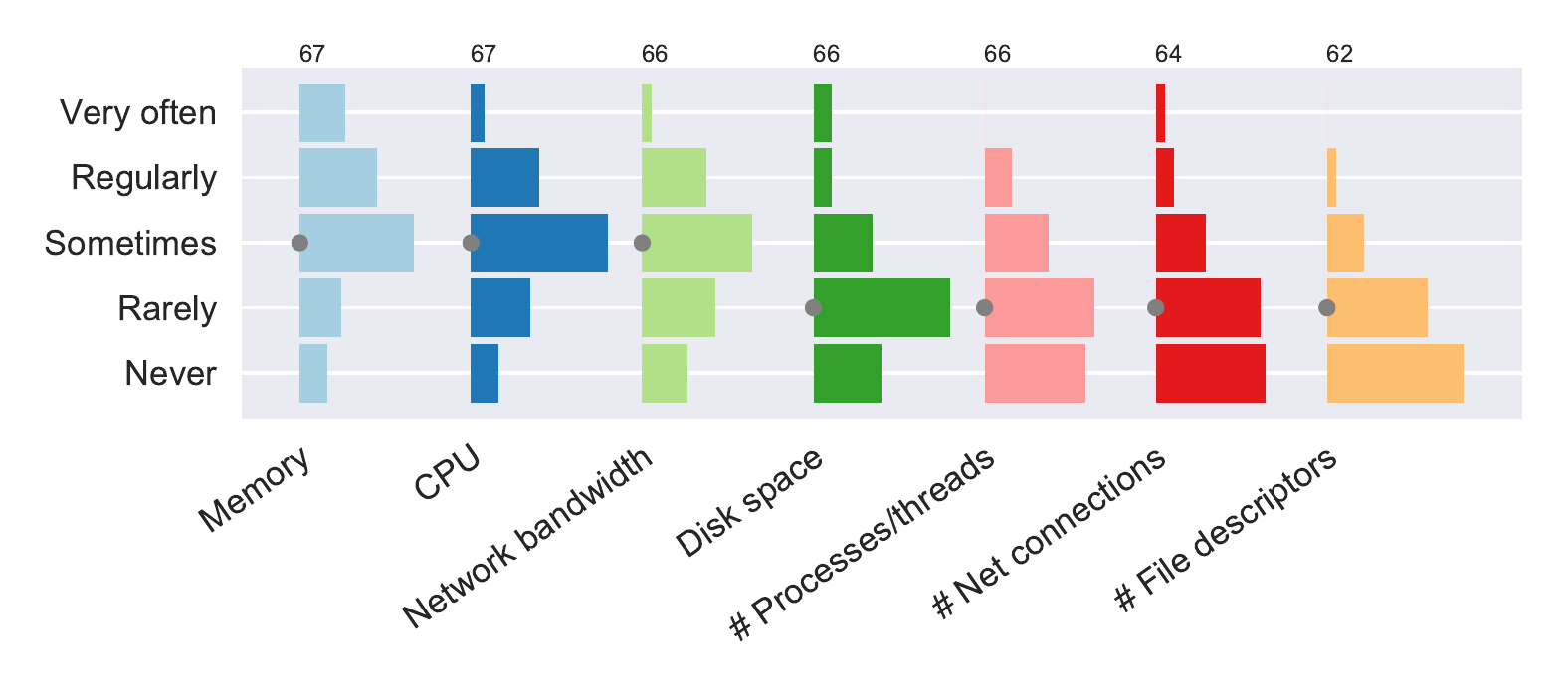}
    \caption[Frequency of bugs effect on system resources]{Frequency of bug effects on system resources.}
    \label{fig:survey-resources}
\end{figure}
Participants also had to rate how frequently different system resources were affected by performance bugs
(\ref{questionnaire:section-89-question-perfbugres}).
These results are visualized in \autoref{fig:survey-resources}.
Memory, CPU and network bandwidth are the most frequently affected resources.
Network bandwidth can be explained by the distributed nature of many of the current robotics systems.
These three primarily affected resources are followed by disk space.
Countable resources like processes or network connections are only rarely affected.
A question for further affected resources (\ref{questionnaire:section-89-question-otherres}) yielded IPC\hyp{}related virtual resources like event queues and IO bandwidth in addition to the previous categories (\cf \autoref{app:summarization_of_other_resources_affected_by_bugs}).

To get an impression of common causes for performance issues in robotics and intelligent systems, a question asked participants to rate how frequently different categories of root causes were the origin of performance bugs in their systems (\ref{questionnaire:section-90-question-perfbugorigin}).
The categories are the ones of the previous general questions on bug origins (\ref{questionnaire:section-90-question-perfbugorigin}) extended with two items specifically targeting performance bugs: skippable computation, \ie unnecessary computation that does not affect the functional outcomes (based on the results in \textcite{Gunawi2014}) and algorithmic choices.
\begin{figure}[t]
    \centering
    \includegraphics[width=0.9\linewidth]{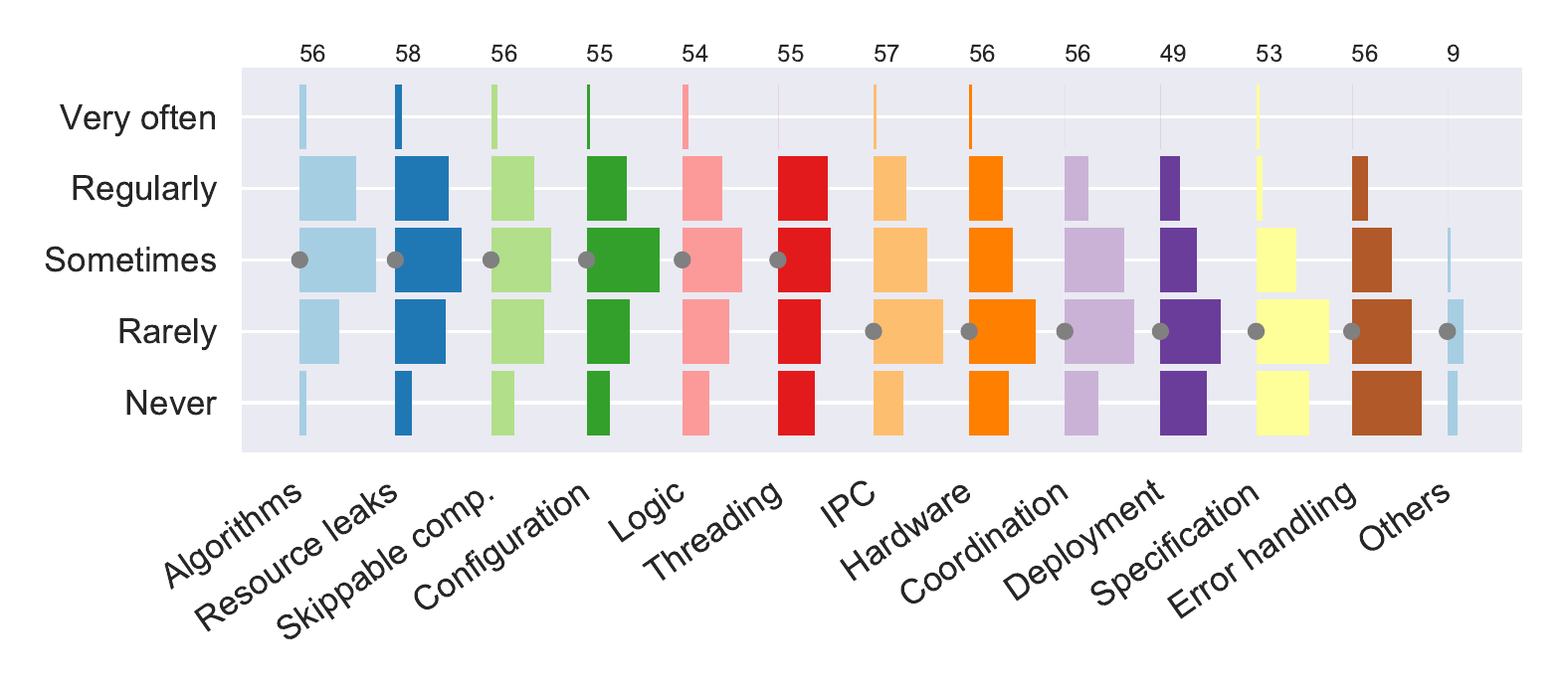}
    \caption[Frequency of reasons for performance bugs]{Frequency of reasons for performance bugs.}
    \label{fig:survey-perfbug-origins}
\end{figure}
\autoref{fig:survey-perfbug-origins} depicts the results for this question.
The most frequent reason for performance bugs is the choice of inappropriate algorithms followed by resource leaks and unnecessary computations.
Interestingly, configuration issues are also among the frequent causes for performance bugs.
When comparing answers to this questions with the answers for origins of general bugs (\ref{questionnaire:section-86-question-bugorigin}), most categories are less likely origins for performance bugs than for general bugs apart from resource leaks (\cf \autoref{tbl:survey-origin-changes}).
Interestingly, threading issues do not significantly affect performance bugs differently than general bugs.
\begin{table}[t]
    \centering
    \begin{tabular}{rS[table-format=-1.2,table-space-text-post={****}]}
        \toprule
        \tableheadline{Category} & \tableheadline{Change} \\
        \midrule
        Communication            &  -0.28  \\
        Configuration            &  -0.51 ** \\
        Coordination             &  -0.79 **** \\
        Deployment               &  -0.71 **** \\
        Error handling           &  -0.44 * \\
        Hardware                 &  -0.98 **** \\
        Logic                    &  -0.44 * \\
        Resource leaks           &   0.47 ** \\
        Specification            &  -0.46 * \\
        Threading                &   0.04  \\
        Others                   &  -0.58  \\
        \bottomrule
    \end{tabular}
    \caption[Failure origins for performance bugs and general bugs]{Changes to the mean ratings for different categories being the origins of failures when comparing performance bugs to general bugs.
        A change of \num{1} would indicate a shift from one answer category to the next higher one.
    Significances have been computed using a Mann\hyp{}Whitney U test.}
    \label{tbl:survey-origin-changes}
\end{table}

\section{Bug examples}
\label{sec:bug_examples}

Finally, participants were asked to provide detailed descriptions of bugs they had observed in their systems.
Two questions in this direction were asked with four sub\hyp{}answers explicitly requesting
\begin{inparaenum}[a)] 
\item the visible effects on the system,
\item the underlying defect causing the bug,
\item the steps performed to debug the problem, and
\item the affected system resources.
\end{inparaenum}
These questions were added to the survey to get an impression of the actual problems current robotics developers are facing with their systems and how they are addressed.

The first of these questions asked for a description of any type of bug that participants remembered from their systems that is particularly representative for the kind of bugs frequently observed (\ref{questionnaire:section-92}).
In total, \num{21} answers were submitted for this question with a complete listing of the answers available in \autoref{app:representative_bugs}.
Most notably, \num{10} of the answers (\SI{48}{\percent}) were related to basic programming issues like segmentation faults or memory leaks, for instance caused by C\fshyp{}C++ peculiarities.
\num{8} answers (\SI{38}{\percent}) described an issue that can be classified as a performance bug.
Issues related to the IPC usage or infrastructure were mentioned by \num{4} answers (\SI{19}{\percent}).
Also, \num{4} answers indicated bugs related to the coordination of the system (for instance, loops in the controlling state machines) of which \num{2} answers were related to unexpected environment situations.
Additionally, \num{2} answers were related to timing aspects and another \num{2} answers indicated that a bug was never or only accidentally understood and solved.
Please refer to the tagging in \autoref{app:representative_bugs} for details.

A second question asked participants to describe the most interesting bug they could remember in the same format.
This was done to get an impression of which extreme types of bugs are possible in robotics systems.
\num{14} participants answered this question and their answers are listed in \autoref{app:interesting_bugs}.
In line with the previous question, programming problems related to low\hyp{}level issues also represent the most frequently mentioned type of bugs with \num{6} answers (\SI{43}{\percent}).
Furthermore, \num{3} answers (\SI{21}{\percent}) described bugs caused by driver or operating system problems.

Answers to both questions indicate that memory\hyp{}management\hyp{}related programming issues are often debugged using established tools like \citesoftware{gdb} or \citesoftware{valgrind} -- however -- with varying success.
One answer specifically mentioned that these tools are often not helpful for distributed systems.

\section{Result interpretation}
\label{sec:result_interpretation}

The presented survey results show that there is still a great potential for improvements in the dependability of robotics systems.
With MTBF rates in the range of hours, a major part of the surveyed systems is far from being reliable enough for longer\hyp{}term operations and work in this direction is needed, even if the majority of developers reached with this survey is working on research systems, which rarely end up in production use cases.
Nevertheless, an appropriate level of dependability is required also in this domain to allow efficient experimentation and reliable studies.
Still, monitoring tools that are specifically geared towards operating systems with high availability and reliability like fault detection or dashboards for a quick manual inspection of systems states are only rarely applied in robotics.
The survey does not provide answers why this is the situation.
Reasons could include the overhead of deploying such approaches which might not be feasible in smaller, relatively short\hyp{}lived systems, or the lack of knowledge about such approaches, especially as many robotics researchers do not have a strong background in maintaining large\hyp{}scale systems.
Therefore, improving approaches and making them more easily usable is one promising direction to foster their application.

With respect to system failures and their origins, the quantitative results from this survey indicate that hardware faults are among the most frequent causes for failure.
This contradicts the findings from \textcite{Steinbauer2013}, which might potentially be caused by the wider range of applications covered in this survey.
Generally, system failures seem to originate more frequently from bugs occurring in high levels of abstraction like coordination, deployment or configuration and less often from component\hyp{}internal issues like crashes.
Still, a majority of the requested bug descriptions for representative bugs actually dealt with such component\hyp{}internal issues.
One reason for this might be that, while frequently being observed, such component\hyp{}related issues are often noticed immediately and therefore are perceived as part of the development work and not as system failures.
In any case, these issues are strikingly often caused by basic programming issues, often related to the manual memory management and syntax idiosyncrasies of C\fshyp{}C++.
A general shift in robotics towards less error\hyp{}prone languages with automatic memory management, clearer syntax and better debugging abilities has the potential to avoid a major amount of bugs currently found during development and operations.

With respect to the performance aspects, one quarter of the bugs found in current systems can be classified as performance bugs.
In the descriptions of representative bugs even more than one third of the answers was performance\hyp{}related.
Therefore, specifically addressing such issues is not only a niche but instead provides the potential to avoid a major amount of failures in the future.
The survey has indicated that performance bugs are significantly less often caused by high\hyp{}level aspects like coordination or deployment and also by hardware issues.
Therefore, addressing them on a component\hyp{}level should already result in reasonable improvements.

Generally, systems are often debugged using log files and \texttt{printf} instructions specifically placed for debugging.
Participants have indicated that debuggers and memory checkers like \citesoftware{valgrind} are used less frequently.
This is probably caused by the fact that such tools cannot be used for all problem kinds.
The detailed bug reports still show that these tools are frequently used to debug programming issues on the component level.
Participants have also indicated that these tools cannot be easily used for problems related to the distributed systems nature of current robots.
Further work on debugging infrastructure respecting this fact might improve the situation.
Finally, simulation seems to be an important tool for debugging robotics systems and explicit support for simulation\hyp{}based testing and debugging might provide one future avenue for more dependable robotics systems.

\section{Threats to validity}
\label{sec:threats_to_validity}

The survey results represent the opinions and memorized impressions of interviewed developers, not objective measurements of the real effects.
As such, results may be biased.
However, general tendencies derived from the results should still be valid as a complete misassessment is unlikely across all participants.

Due to the distribution of the survey via primarily research\hyp{}related mailing lists, results are only representative for systems developed in this context and cannot be generalized towards industrial, production\hyp{}ready systems.

The categories used in questions regarding the frequencies of bug origins may have partially been hard to distinguish from each other.
Therefore, in some cases, ratings might be blurred between multiple categories due to the imprecise definitions.
When possible, the conclusions drawn from the survey have been based on a grouping of multiple categories to mitigate this effect.

\appendix

\section{Questionnaire structure}
\label{app:survey}

The following sections represent the structure of the online survey.
This is a direct export of the survey structure without modifications.

\subsection{Introduction}
\label{questionnaire:introduction}

	Thank you very much for taking the time to participate in this survey. This survey is part of my PhD project with a focus on exploiting knowledge about computational resource consumption in robotics and intelligent systems, persued at Bielefeld University. Therefore, in order to participate, you should be involved or have been involved in the development or maintenance of such systems. In case you have worked or are working with mutiple systems in parallel, please provide answers on the combination of all theses systems.

	Participating in this survey should not take longer than 15 minutes. The survey consists of several questions and you are free to skip questions in case you do not want to answer them. Moreover, you can go back and forth between the questions you have already answered in order to revise them. All data you enter in this survey will be anonymized.

	Johannes Wienke

	jwienke [at] techfak.uni-bielefeld.de

\subsection{Monitoring Tools}
\label{questionnaire:section-88}

	The first part of this survey addresses how robotics and intelligent systems are monitored at runtime in order to assess their health and understand the ongoing operations. Monitoring includes the ongoing collection of runtime data, the observation of operations as well as the assessment of system health.

\subsubsection{How often do you use the following kinds of tools to monitor the operation of running systems? }
\label{questionnaire:section-88-question-monitoringkinds}

Rate individually for:
\begin{itemize}
    \item Operating system command line tools
e.g.\ htop, iotop, ps (OS)
    \item Logfiles (LOG)
    \item Dashboard views
e.g.\ munin, graphite, nagios (DASH)
    \item Inter-process communication introspection
e.g.\ middleware logger (IPC)
    \item Autonomous fault or anomaly detectors (FD)
    \item Special-purpose visualizations
e.g.\ rviz, image processing debug windows (VIS)
    \item Remote desktop connections
e.g.\ VNC, rdesktop (RDP)
    \item Others (OTH)
\end{itemize}

\paragraph{Answer type}
Fixed choice
\begin{itemize}
    \item Never (0)
    \item Rarely (1)
    \item Sometimes (2)
    \item Regularly (3)
    \item Always (4)
\end{itemize}

\subsubsection{Please name the concrete tools that you use for monitoring running systems.}
\label{questionnaire:section-88-question-monitoringtools}

Separate different tools with a comma.

\paragraph{Answer type}
longtext (length: 40)

\subsection{Debugging Tools}
\label{questionnaire:section-85}

This part of the survey addresses tools that are used in order to debug systems in case a failure has been detected. Debugging is the process of identifying the root cause of an observed abnormal system behavior.

\subsubsection{How often do you use the following tools for debugging?}
\label{questionnaire:section-85-question-debuggingkinds}

Rate individually for:
\begin{itemize}
    \item Console output
e.g.\ printf, cout (PRNT)
    \item Logfiles (LOG)
    \item Debuggers
e.g.\ gdb, pdb (DBG)
    \item Profilers
e.g.\ kcachegrind, callgrind (PROF)
    \item Memory checkers
e.g.\ valgrind (MEMC)
    \item System call introspection
e.g.\ strace, systemtap (SYSC)
    \item Inter-process communication introspection
e.g.\ middleware logger (IPC)
    \item Network analyzers
e.g.\ wireshark (NWAN)
    \item Automated testing
e.g.\ unit tests (TEST)
    \item Simulation (SIM)
    \item Others (OTH)
\end{itemize}

\paragraph{Answer type}
Fixed choice
\begin{itemize}
    \item Never (0)
    \item Rarely (1)
    \item Sometimes (2)
    \item Regularly (3)
    \item Always (4)
\end{itemize}

\subsubsection{Please name the concrete tools that you use for debugging.}
\label{questionnaire:section-85-question-debuggingtools}

Separate different tools with a comma.

\paragraph{Answer type}
longtext (length: 40)

\subsection{General Failure Assessment}
\label{questionnaire:section-86}

Please provide information about failures you have observed in the systems you are working with.

\subsubsection{Averaging over the systems you have been working with, what to do you think is the mean time between failures for these systems?}
\label{questionnaire:section-86-question-mtbf}

The mean time between failures is the average amount of operation time of a system until a failure occurs.

\paragraph{Answer type}
Fixed choice
\begin{itemize}
    \item < 0.5 hours (0)
    \item < 1 hour (1)
    \item < 6 hours (2)
    \item < 12 hours (3)
    \item < 1 week (4)
    \item > 1 week (5)
\end{itemize}

\subsubsection{Please indicate how often the following items were the root cause for system failures that you know about.}
\label{questionnaire:section-86-question-bugorigin}

Rate individually for:
\begin{itemize}
    \item Hardware issues (HW)
    \item System coordination
e.g.\ state machine (COORD)
    \item Deployment (DEPL)
    \item Configuration errors
e.g.\ component configuration (CONF)
    \item Logic errors (LOGIC)
    \item Threading and synchronization (THRD)
    \item Wrong error handling code (ERR)
    \item Resource leaks or starvation
e.g.\ RAM full, CPU overloaded (LEAK)
    \item Inter-process communication failures
e.g.\ dropped connection, protocol error (COMM)
    \item Specification error / mismatch
e.g.\ component receives other inputs than specified (SPEC)
    \item Others (OTH)
\end{itemize}

\paragraph{Answer type}
Fixed choice
\begin{itemize}
    \item Never (0)
    \item Rarely (1)
    \item Sometimes (2)
    \item Regularly (3)
    \item Very often (4)
\end{itemize}

\subsubsection{Which other classes of root causes for failures did you observe? }
\label{questionnaire:section-86-question-bugoriginother}

Separate items by comma.

\paragraph{Answer type}
text (length: 24)

\subsection{Resource-Related Bugs}
\label{questionnaire:section-87}

The following questions deal with the consumption of computational resources like CPU, memory, disk, network etc.

\subsubsection{How many of the bugs you have observed or know about had an impact on computational resources,  e.g.\ by consuming more or less of these resources as expected?}
\label{questionnaire:section-87-question-perfbugrate}

Please approximate the amount with a percentage value of the total number of bugs you can remember. A quick guess is ok here.

\paragraph{Answer type}
integer (length: 10)

\subsection{Impact on Computational Resources}
\label{questionnaire:section-89}

The following questions deal with the consumption of computational resources like CPU, memory, disk, network etc.

\subsubsection{Please indicate how often the following computational resources were affected by bugs you have observed.}
\label{questionnaire:section-89-question-perfbugres}

A computational resource was affected by a bug in case its consumption was higher or less than expected, e.g.\ in comparable or non-faulty situations.

Rate individually for:
\begin{itemize}
    \item CPU (CPU)
    \item Working memory (MEM)
    \item Hard disc space (HDD)
    \item Network bandwidth (NET)
    \item Number of network connections (CON)
    \item Number of processes and threads (PROC)
    \item Number of file descriptors (DESC)
\end{itemize}

\paragraph{Answer type}
Fixed choice
\begin{itemize}
    \item Never (0)
    \item Rarely (1)
    \item Sometimes (2)
    \item Regularly (3)
    \item Very often (4)
\end{itemize}

\subsubsection{If there are other computational resources that have been affected by bugs, please name these.}
\label{questionnaire:section-89-question-otherres}

\paragraph{Answer type}
longtext (length: 40)

\subsection{Performance Bugs}
\label{questionnaire:section-90}

The following question specifically addresses performance bugs. A system failure or bug is a performance bug in case it is visible either through degradation in the observed performance of the system (e.g.\ delayed or very slow reactions) or through an unexpected consumption of computational resources like CPU, memory, disk, network etc.

\subsubsection{Please rate how often the following items were the root causes for performance bugs you have observed.}
\label{questionnaire:section-90-question-perfbugorigin}

Rate individually for:
\begin{itemize}
    \item Hardware issues (HW)
    \item System coordination
e.g.\ state machine (COORD)
    \item Deployment (DEPL)
    \item Configuration errors
e.g.\ component configuration (CONF)
    \item Logic errors (LOGIC)
    \item Threading and synchronization (THRD)
    \item Wrong error handling code (ERR)
    \item Unnecessary or skippable computation (SKIP)
    \item Resource leaks or starvation
e.g.\ RAM full, CPU overloaded (LEAK)
    \item Inter-process communication failures
e.g.\ dropped connection, protocol error (COMM)
    \item Specification error / mismatch (SPEC)
    \item Algorithm choice (ALGO)
    \item Others (OTH)
\end{itemize}

\paragraph{Answer type}
Fixed choice
\begin{itemize}
    \item Never (0)
    \item Rarely (1)
    \item Sometimes (2)
    \item Regularly (3)
    \item Always (4)
\end{itemize}

\subsection{Case Studies}
\label{questionnaire:section-95}

For the following questions, please provide descriptions of any kind of bug that you remember.

\subsubsection{Thinking about the systems you have worked with so far, is there a bug that you remember which happened several times or which is representative for a class of comparable bugs?}
\label{questionnaire:section-95-question-repbug}

\paragraph{Answer type}
Fixed choice
\begin{itemize}
    \item Yes (Y)
    \item No (N)
\end{itemize}

\subsection{Case Study: Representative Bug}
\label{questionnaire:section-92}

Please briefly describe the representative bug that you remember.

\subsubsection{How was the representative bug noticed?}
\label{questionnaire:section-92-question-observation}

Please explain the observations that were made and how they diverged from the expectations.

\paragraph{Answer type}
longtext (length: 40)

\subsubsection{What was the root cause for the bug?}
\label{questionnaire:section-92-question-origin}

Please explain which component(s) of the system failed and in which way.

\paragraph{Answer type}
longtext (length: 40)

\subsubsection{Which steps were necessary to analyze and debug the problem?}
\label{questionnaire:section-92-question-debugging}

Please include the information sources that had to be observed and the tools that got applied.

\paragraph{Answer type}
longtext (length: 40)

\subsubsection{Which computational resources were affected by the bug?}
\label{questionnaire:section-92-question-resources}

Computational resources include CPU, working memory, hard disc space, network bandwidth \& connections, number of processes and threads, nubmer of file descriptors etc.

\paragraph{Answer type}
longtext (length: 40)

\subsection{Case Studies}
\label{questionnaire:section-96}

For the following questions, please describe any kind of bug that you remember.

\subsubsection{Thinking about the systems you have worked with so far, is there a bug that you remember which was particularly interesting for you?}
\label{questionnaire:section-96-question-intbug}

\paragraph{Answer type}
Fixed choice
\begin{itemize}
    \item Yes (Y)
    \item No (N)
\end{itemize}

\subsection{Case Study: Interesting Bug}
\label{questionnaire:section-93}

Please describe briefly the most interesting bug that you remember from one of the systems you have been working with.

\subsubsection{How was the interesting bug noticed?}
\label{questionnaire:section-93-question-observation}

Please explain the observations that were made and how they diverged from the expectations.

\paragraph{Answer type}
longtext (length: 40)

\subsubsection{What was the root cause for the bug?}
\label{questionnaire:section-93-question-origin}

Please explain which component(s) of the system failed and in which way.

\paragraph{Answer type}
longtext (length: 40)

\subsubsection{Which steps were necessary to analyze and debug the problem?}
\label{questionnaire:section-93-question-debugging}

Please include the information sources that had to be observed and the tools that got applied.

\paragraph{Answer type}
longtext (length: 40)

\subsubsection{Which computational resources were affected by the bug?}
\label{questionnaire:section-93-question-resources}

Computational resources include CPU, working memory, hard disc space, network bandwidth \& connections, number of processes and threads, nubmer of file descriptors etc.

\paragraph{Answer type}
longtext (length: 40)

\subsection{Personal Information}
\label{questionnaire:section-94}

As a final step, please provide some information about your experience with robotics and intelligent systems development.

\subsubsection{In which context do you develop robotics or intelligent systems?}
\label{questionnaire:section-94-question-context}

\paragraph{Answer type}
Fixed choice
\begin{itemize}
    \item Student (excluding PhD students) (STUD)
    \item Researcher at a university (PhD students, scientific staff) (RES)
    \item Industry (IND)
    \item Other (OTHER)
\end{itemize}

\subsubsection{How many years of experience in robotics and intelligent systems development do you have?}
\label{questionnaire:section-94-question-experience}

\paragraph{Answer type}
integer (length: 10)

\subsubsection{How much of your time do you spend on developing in the following domains?}
\label{questionnaire:section-94-question-perc}

Please indicate in percent of total development time. Numbers may not sum up to 100.

Rate individually for:
\begin{itemize}
    \item Hardware (HW)
    \item Drivers (DRV)
    \item Functional components (COMP)
    \item Inter-process communication infrastructure (COMM)
    \item Software architecture and integration (ARCH)
    \item Other (ANY)
\end{itemize}

\paragraph{Answer type}
integer (length: 3)
Hint: Percent of development time

\subsection{Final remarks}
\label{questionnaire:final-remarks}

	Thank you very much for participating in this survey and thereby supporting my research.

	In case you have further questions regarding this survey or the research topic in general, please contact me via email.

	Johannes Wienke

	jwienke [at] techfak.uni-bielefeld.de

\section{Result details}
\label{app:survey-results}

\subsection{Used monitoring tools}
\label{app:used_monitoring_tools}

The following table presents the results for question~\ref{questionnaire:section-88-question-monitoringtools}.
The free text answers have been been grouped into categories (caption lines in the table).
For each answer that included at least one item belonging to a category, the counter of each category was incremented.
Hence, the counts represent the number of answers that mentioned a category at least once.
Additionally, for each category, representative entries have been counted the same way.
Some of the answers include uncommon or special-purpose tools or techniques.
These have not been counted individually and, hence, are only visible in the category counts.

\begingroup
\setlength\extrarowheight{0pt}
\noindent\begin{longtable}[c]{l *{2}{S[table-format=2.0,table-column-width=1.7cm]}}
    \toprule
    \tableheadline{Tool}                          & \multicolumn{2}{c}{\spacedlowsmallcaps{Answer count}} \\
    \midrule\endhead
    \midrule
    \multicolumn{3}{r}{\textit{\scriptsize Continued on next page}} \\
    \endfoot
    \bottomrule
    \endlastfoot
    \spacedlowsmallcaps{Visualization}                         & 27 &    \\*
    {\scriptsize\hspace{8mm}rviz}                              &    & {\scriptsize 22} \\*
    {\scriptsize\hspace{8mm}gnuplot}                           &    & {\scriptsize 2}  \\*
    {\scriptsize\hspace{8mm}matplotlib}                        &    & {\scriptsize 1}  \\
    \spacedlowsmallcaps{Middleware Tools}\protect\footnote{Represents entries that are specific to the middleware-related aspects of an ecosystem. For instance, \texttt{ROS\_DEBUG} has not been counted here.  Instead, this belongs to the \textquote{Manual log reading} category.}                      & 23 &    \\*
    {\scriptsize\hspace{8mm}ROS command line}                  &    & {\scriptsize 14} \\*
    {\scriptsize\hspace{8mm}rqt}                               &    & {\scriptsize 5}  \\*
    {\scriptsize\hspace{8mm}RSB}                               &    & {\scriptsize 4}  \\
    \spacedlowsmallcaps{Basic OS Tools}                        & 22 &    \\*
    {\scriptsize\hspace{8mm}htop}                              &    & {\scriptsize 12} \\*
    {\scriptsize\hspace{8mm}ps}                                &    & {\scriptsize 7}  \\*
    {\scriptsize\hspace{8mm}top}                               &    & {\scriptsize 7}  \\*
    {\scriptsize\hspace{8mm}acpi}                              &    & {\scriptsize 1}  \\*
    {\scriptsize\hspace{8mm}du}                                &    & {\scriptsize 1}  \\*
    {\scriptsize\hspace{8mm}free}                              &    & {\scriptsize 1}  \\*
    {\scriptsize\hspace{8mm}lsof}                              &    & {\scriptsize 1}  \\*
    {\scriptsize\hspace{8mm}procman (gnome)}                   &    & {\scriptsize 1}  \\*
    {\scriptsize\hspace{8mm}pstree}                            &    & {\scriptsize 1}  \\*
    {\scriptsize\hspace{8mm}screen}                            &    & {\scriptsize 1}  \\*
    {\scriptsize\hspace{8mm}tmux}                              &    & {\scriptsize 1}  \\
    \spacedlowsmallcaps{Manual Log Reading}                    & 13 &    \\
    \spacedlowsmallcaps{Remote Access}                         & 9  &    \\*
    {\scriptsize\hspace{8mm}ssh}                               &    & {\scriptsize 5}  \\*
    {\scriptsize\hspace{8mm}putty}                             &    & {\scriptsize 1}  \\*
    {\scriptsize\hspace{8mm}rdesktop}                          &    & {\scriptsize 1}  \\*
    {\scriptsize\hspace{8mm}vnc}                               &    & {\scriptsize 1}  \\
    \spacedlowsmallcaps{Custom Mission-Specific}               & 4  &    \\
    \spacedlowsmallcaps{Generic Network}                       & 2  &    \\*
    {\scriptsize\hspace{8mm}netstat}                           &    & {\scriptsize 1}  \\*
    {\scriptsize\hspace{8mm}tcpdump}                           &    & {\scriptsize 1}  \\*
    {\scriptsize\hspace{8mm}wireshark}                         &    & {\scriptsize 1}  \\
    \spacedlowsmallcaps{Hardware Signals}                      & 1  &    \\
\end{longtable}
\endgroup

\subsection{Used debugging tools}
\label{app:used_debugging_tools}

The following table presents the results for question~\ref{questionnaire:section-85-question-debuggingtools}.
The free text answers have been been grouped into categories (highlighted lines in the table).
For each answer that included at least one item belonging to a category, the counter of each category was incremented.
Hence, the counts represent the number of answers that mentioned a category at least once.
Additionally, for each category, representative entries have been counted the same way.
Some of the answers include uncommon or special-purpose tools or techniques.
These have not been counted individually and, hence, are only visible in the category counts.

\begingroup
\setlength\extrarowheight{0pt}
\noindent\begin{longtable}[c]{l *{2}{S[table-format=2.0,table-column-width=1.7cm]}}
    \toprule
    \tableheadline{Tool}                          & \multicolumn{2}{c}{\spacedlowsmallcaps{Answer count}} \\
    \midrule\endhead
    \midrule
    \multicolumn{3}{r}{\textit{\scriptsize Continued on next page}} \\
    \endfoot
    \bottomrule
    \endlastfoot
    \spacedlowsmallcaps{Debuggers}                   & 19 &    \\*
    {\scriptsize\hspace{8mm}gdb}                     &    & {\scriptsize 17} \\*
    {\scriptsize\hspace{8mm}pdb}                     &    & {\scriptsize 3}  \\*
    {\scriptsize\hspace{8mm}VS debugger}             &    & {\scriptsize 2}  \\*
    {\scriptsize\hspace{8mm}ddd}                     &    & {\scriptsize 1}  \\*
    {\scriptsize\hspace{8mm}jdb}                     &    & {\scriptsize 1}  \\
    \spacedlowsmallcaps{Runtime Intropsection}            & 13 &    \\*
    {\scriptsize\hspace{8mm}valgrind}                &    & {\scriptsize 12} \\*
    {\scriptsize\hspace{8mm}callgrind}               &    & {\scriptsize 2}  \\*
    {\scriptsize\hspace{8mm}kcachegrind}             &    & {\scriptsize 1}  \\*
    {\scriptsize\hspace{8mm}strace}                  &    & {\scriptsize 1}  \\
    \spacedlowsmallcaps{Generic}                     & 15 &    \\*
    {\scriptsize\hspace{8mm}printf, cout, \etc}      &    & {\scriptsize 14} \\*
    {\scriptsize\hspace{8mm}logfiles}                &    & {\scriptsize 4}  \\*
    {\scriptsize\hspace{8mm}git}                     &    & {\scriptsize 1}  \\
    \spacedlowsmallcaps{Middleware Tools}\protect\footnote{Represents entries that are specific to the middleware-related aspects of an ecosystem.}                  & 12 &    \\*
    {\scriptsize\hspace{8mm}ROS command line}        &    & {\scriptsize 5}  \\*
    {\scriptsize\hspace{8mm}RQT}                     &    & {\scriptsize 2}  \\*
    {\scriptsize\hspace{8mm}RSB}                     &    & {\scriptsize 2}  \\
    \spacedlowsmallcaps{Simulation \& Visualization} & 7  &    \\*
    {\scriptsize\hspace{8mm}gazebo}                  &    & {\scriptsize 4}  \\*
    {\scriptsize\hspace{8mm}rviz}                    &    & {\scriptsize 1}  \\*
    {\scriptsize\hspace{8mm}Vortex}                  &    & {\scriptsize 1}  \\*
    {\scriptsize\hspace{8mm}stage}                   &    & {\scriptsize 1}  \\
    \spacedlowsmallcaps{Functional Testing}          & 6  &    \\*
    {\scriptsize\hspace{8mm}gtest}                   &    & {\scriptsize 2}  \\*
    {\scriptsize\hspace{8mm}junit}                   &    & {\scriptsize 2}  \\*
    {\scriptsize\hspace{8mm}cppunit}                 &    & {\scriptsize 1}  \\*
    {\scriptsize\hspace{8mm}rostest}                 &    & {\scriptsize 1}  \\
    \spacedlowsmallcaps{IDEs}                        & 4  &    \\*
    {\scriptsize\hspace{8mm}Qt Creator}              &    & {\scriptsize 2}  \\*
    {\scriptsize\hspace{8mm}KDevelop}                &    & {\scriptsize 1}  \\*
    {\scriptsize\hspace{8mm}LabVIEW}                 &    & {\scriptsize 1}  \\*
    {\scriptsize\hspace{8mm}Matlab}                  &    & {\scriptsize 1}  \\*
    {\scriptsize\hspace{8mm}Visual Studio}           &    & {\scriptsize 1}  \\
    \spacedlowsmallcaps{Generic Network}             & 2  &    \\*
    {\scriptsize\hspace{8mm}wireshark}               &    & {\scriptsize 2}  \\*
    {\scriptsize\hspace{8mm}tcpdump}                 &    & {\scriptsize 1}  \\
    \spacedlowsmallcaps{Dynamic Analysis}            & 1  &    \\*
    {\scriptsize\hspace{8mm}Daikon}                  &    & {\scriptsize 1}  \\
    \end{longtable}
\endgroup

\subsection{Summarization of free form bug origins}
\label{app:summarization_of_free_form_bug_origins}

The following table presents all answers to question~\ref{questionnaire:section-86-question-bugoriginother}.
Individual answers have been split into distinct aspects.
These aspects have either been assigned to an existing answer category from question~\ref{questionnaire:section-86-question-bugorigin} or to new categories.

\bigskip

\begingroup
\footnotesize
\setlength\LTleft{0pt}
\setlength\LTright{0pt}
\begin{longtable}{@{\extracolsep{\fill}}p{6cm}cc}
    \toprule
    \tableheadline{Answer} & \multicolumn{2}{c}{\spacedlowsmallcaps{Category}} \\
    & \tableheadline{Existing} & \tableheadline{New} \\
    \midrule\endhead
    \midrule
    \multicolumn{3}{r}{\textit{\scriptsize Continued on next page}} \\
    \endfoot
    \bottomrule
    \endlastfoot
    unknown driver init problems (start a driver, and works only after second trial) & & Driver \& OS \\*
    environment noise (lighting condition variation, sound condition in speach recognition) hard to adapt to every possible variation & & Environment \\
    \addlinespace[0.61cm]
    Insufficient Component Specifications & Specification & \\
    \addlinespace[0.61cm]
    Changed maps\fshyp{}environments & & Environment \\*
    lossy WiFi connections & Hardware & \\*
    unreliable hardware & Hardware & \\
    \addlinespace[0.61cm]
    in Field robotics, the environment is the first enemy\dots & & Environment \\
    \addlinespace[0.61cm]
    Environment changes & & Environment \\
    sensor failures & Hardware & \\
    \addlinespace[0.61cm]
    unprofessional users & & Environment \\
    \addlinespace[0.61cm]
    Operation System / Firmware failure & & Driver \& OS \\
    \addlinespace[0.61cm]
    network too slow & Hardware & \\
    \addlinespace[0.61cm]
    Loose wires & Hardware & \\*
    other researchers changing the robot configuration & & Config mgmt \\
    \addlinespace[0.61cm]
    coding bugs & Logic & \\
    \addlinespace[0.61cm]
    algorithm limitations & & Environment \\*
    sensor limitations & & Hardware lim \\*
    perception limitations & & Environment \\
    \addlinespace[0.61cm]
    wrong usage & & Environment \\
    \addlinespace[0.61cm]
    Failures in RT OS timing guarantees & & Driver \& OS \\
\end{longtable}
\endgroup

\subsection{Summarization of other resources affected by bugs}
\label{app:summarization_of_other_resources_affected_by_bugs}

The following table presents the free text results of question~\ref{questionnaire:section-89-question-otherres}.
Answers have been split into distinct aspects and these aspects have either been assigned to one of the existing categories from
question~\ref{questionnaire:section-89-question-perfbugres} or -- if these did not match -- new categories have been created to capture the answers.
Parts of answers that did not represent system resources which have a resource capacity  that can be utilized have been ignored.
These are marked as strikethrough text.

\bigskip

\begingroup
\setlength\LTleft{0pt}
\setlength\LTright{0pt}
\begin{longtable}{@{\extracolsep{\fill}}p{5.3cm}cc}
    \toprule
    \tableheadline{Answer} &
    \multicolumn{2}{c}{\spacedlowsmallcaps{Resource}} \\
    & \tableheadline{Existing} & \tableheadline{New} \\
    \midrule\endhead
    \midrule
    \multicolumn{3}{r}{\textit{\scriptsize Continued on next page}} \\
    \endfoot
    \bottomrule
    \endlastfoot
    USB bandwidth and or stability & & IO bandwidth \\
    \addlinespace[0.61cm]
    locks on files\fshyp{}devices\fshyp{}resources & File descriptors & \\
    \st{permissions} & & \\
    \st{file system integrity} & & \\
    \addlinespace[0.61cm]
    interprocess communication queues, \eg queue overflow & & IPC \\
    \addlinespace[0.61cm]
    Files (devices) left open. & File descriptors & \\
    \st{Wrong operation in GPU leads to restart.} & & \\
    \addlinespace[0.61cm]
    Memory leak -- not sure why or where & Memory & \\
\end{longtable}
\endgroup

\subsection{Representative bugs}
\label{app:representative_bugs}

The following subsections present answers to the questions for representative bugs (\ref{questionnaire:section-92}).
For the analysis, answers have been tagged for various aspects and types of bugs being mentioned in them.
Raw submission texts have been reformatted to match the document and typographical and grammatical errors have been corrected.

\begingroup
\setlength{\emergencystretch}{3em}

\newcommand{\bugdesc}[6]{%
    \subsubsection{Representativ bug #1}%
    \label{app:representative_bugs:#1}%
    \paragraph{Observation} #2%
    \paragraph{Cause} #3%
    \paragraph{Debugging} #4%
    \paragraph{Affected resources} #5%
    \paragraph{Tags} #6}

\bugdesc%
{8}%
{computer system unresponsive}%
{memory leak}%
{\begin{itemize}
    \item find faulty process
    \item analyze memory usage (\texttt{valgrind}\fshyp\texttt{gdb})
    \item repair code
\end{itemize}}%
{main memory}%
{basic programming issue; performance bug}

\bugdesc%
{10}%
{System got stuck in infinite loop.}%
{Unexpected infinite loop in the behaviour (state machine).
Noise in the data caused the system to infinitely switch between two states.}%
{\begin{enumerate}
    \item Detection of which states were affected.
    \item Detection of the responsible subsystem(s). 
    \item Detection of the responsible functions.
    \item Recording data that caused the problem.
    \item Analyzing the data and searching for unexpected situations.
    \item Modification of the system in order to handle such situation correctly.
\end{enumerate}}%
{CPU}%
{coordination; environment-related}

\bugdesc%
{14}%
{high latency in spread communication}%
{wrong spread configuration\fshyp wrong deployment of components}%
{trial \& error: reconfiguration, stopping and starting components, monitoring of latency via \texttt{rsb-tools}}%
{network-latency}%
{communication; performance bug}

\bugdesc%
{21}%
{Incorrect response of the overall system according to requested task request.
System thinks it did not grasp an object although it did and restarts grasping operation or cancels the task due to the missing object in hand.}%
{State machine design and\fshyp or logic error and\fshyp or untriggered event due to sensor not triggering as expected (hardware) or too much noise (environment noise).
The root cause is often a case not being handled correctly in a big system with a lot of sensors and possible case.}%
{event logger analysis over XCF XML data, unit test of single sensor output to see noise level or false positives.}%
{Hardware (noise in the sensor)}%
{coordination; environment-related}

\bugdesc%
{26}%
{Segfault}%
{Segfault}%
{\texttt{gdb}}%
{}%
{basic programming issue}

\bugdesc%
{30}%
{Unexpected overall behavior.}%
{Wrong logic in the abstract level.}%
{Run simulation in the abstract layer.}%
{None.}%
{coordination}

\bugdesc%
{41}%
{Failure to observe expected high-level output.
More specifically, a map that was being built was lacking data.}%
{Congested wireless network connection.
The amount of data could not be transmitted within the expected time frame.}%
{Logging of signals between modules on the deployed system to verify data was being produced and transmitted correctly, and logging of data received.}%
{Network connection}%
{communication; timing}

\bugdesc%
{42}%
{Because of timing mismatch the planning system was working with outdated data.}%
{Non-event based data transfer.}%
{Going through multiple log files in parallel to find the data that was transmitted in comparison to the data that was used in the computation.}%
{Non. Mostly mismatch between specification and performed actions.}%
{coordination; timing}

\bugdesc%
{46}%
{Navigation did not work correctly}%
{Algorithmic errors}%
{Dig in and verify steps in the algorithm}%
{}%
{}

\bugdesc%
{60}%
{delays in robots command execution}%
{supervision and management part of the framework}%
{benchmarking, profiling}%
{}%
{performance bug}

\bugdesc%
{69}%
{memory leak}%
{resource management, dangling pointers}%
{check, object\fshyp resource timeline, usually start with resources that are created often and handed over regularly and therefore might have unclear ownership}%
{memory, CPU}%
{basic programming issue; performance bug}

\bugdesc%
{70}%
{constantly increasing memory consumption}%
{Memory leaks}%
{Running the code in offline mode with externally provided inputs and observing the memory consumption pattern. Tools like \texttt{valgrind} or system process monitor helps to discover the problem}%
{Working memory}%
{basic programming issue; performance bug}

\bugdesc%
{76}%
{Visually in system operation.
In one case, elements within a graphical display were misdrawn. In another, command codes were misinterpreted, resulting in incorrect system operation.}%
{Variable type mismatch \eg integer vs.\ unsigned integer -- such as when a number intended to be a signed integer is interpreted as an unsigned integer by another subsystem.}%
{Debugger using single step and memory access.}%
{None}%
{basic programming issue; performance bug}

\bugdesc%
{81}%
{segfault}%
{C++ pointers}%
{\texttt{gdb}, \texttt{valgrind}}%
{none}%
{basic programming issue}

\bugdesc%
{96}%
{segmentation fault}%
{logical errors, bad memory management}%
{using debuggers, looking and studying code}%
{working memory, number of process and threads}%
{basic programming issue}

\bugdesc%
{128}%
{Robot software is not working / partially working (\eg recognizing and grasping an object)}%
{Wrong configuration and\fshyp or API changes that hasn't been changes in all components (Problem with scripting languages like python)}%
{\begin{itemize}
    \item identify error message and component via log files / console output
    \item Think about what could have caused the problem (look into source code, \texttt{git}\fshyp\texttt{svn} commit messages\fshyp diffs)
    \item try to fix it directly or talk with other developers in case of bigger changes / out of my responsibility
\end{itemize}}%
{none}%
{}

\bugdesc%
{135}%
{middleware communication stopped / was only available within small subsets of components}%
{unknown}%
{}%
{}%
{not\fshyp{}accidentally solved; communication}

\bugdesc%
{136}%
{\begin{enumerate}
    \item Application\fshyp process hang.
    \item \SI{100}{\percent} core usage on idle
    \item Unbalanced load between cores (Monolithic code).
\end{enumerate}}%
{\begin{enumerate}
    \item Loose wire\fshyp couple (mostly USB)
    \item Active wait\newline
        \texttt{while(1) { while(!flag); process(); flag = 0; }}
    \item A bad design. No threads were used, but time measurements to switch between tasks.
\end{enumerate}}%
{\begin{enumerate}
    \item Check everything, realize that the file-device is open but device is no longer present or has different pointer or has reseted
    \item[2/3.] Check every code file. People use to make old-style structured programming when using C/C++ 
\end{enumerate}
when you notice the performance go brick, check CPU\fshyp memory usage with OS tools and notice one process is using everything but is idle.}%
{Mostly CPU}%
{basic programming issue; performance bug}

\bugdesc%
{156}%
{Difficult to reproduce, random segmentation faults}%
{\SI{90}{\percent} of the time it has been either accessing unallocated memory (off-by-one errors) or threading issues}%
{When working with a system with many processes, threads, inter-process communications, \etc, the standard tools (\texttt{gdb}, \texttt{valgrind}) are often not that helpful.
If they can't immediately point me to the error, I'll often resort to print statement debugging.}%
{Memory leaks, CPU usage}%
{basic programming issue}

\bugdesc%
{190}%
{unforeseen system behavior, decreased system performance}%
{misconfiguration of middleware}%
{\begin{itemize}
    \item monitoring middleware configuration of concerned components
    \item checking log-files
    \item sometimes debug print-outs
\end{itemize}}%
{CPU, network load}%
{communication; performance bug}

\bugdesc%
{191}%
{Software controlling the robot crashed immediately after started in robot or robot stop to move when has to perform certain operation}%
{The error was caused by not checking range of allocated memory in some object's constructor, we used \texttt{sprintf} instead of \texttt{snprintf}}%
{\begin{itemize}
    \item \texttt{gdb} -- did not find anything
    \item \texttt{valgrind} -- did not find anything
\end{itemize}
Both tools were run on PC, where the error did not occur, but we did not use them on the robot's pc. The bug was found accidentally.}%
{access to non-allocated memory lead immediately to crash of program.}%
{basic programming issue; not\fshyp{}accidentally solved}

\endgroup

\subsection{Interesting bugs}
\label{app:interesting_bugs}

The following subsections present answers to the questions for intersting bugs (\ref{questionnaire:section-93}).
Answers have been processed the same way as for \autoref{app:representative_bugs}.

\begingroup
\setlength{\emergencystretch}{3em}

\newcommand{\bugdesc}[6]{%
    \subsubsection{Interesting bug #1}%
    \label{app:interesting_bugs:#1}%
    \paragraph{Observation} #2%
    \paragraph{Cause} #3%
    \paragraph{Debugging} #4%
    \paragraph{Affected resources} #5%
    \paragraph{Tags} #6}

\bugdesc%
{5}%
{There are too many to remember.
A recent one got noticed by surprisingly high latency in a multithreaded processing and visualization pipeline.}%
{Sync to vblank was enabled on a system and due to a possible bug in Qt multiple GL widgets contributed to the update frequency.
The maximum display update frequency dropped below \SI{30}{\hertz}.}%
{Compare systems and analyze timing inside the application. Google the problem.}%
{None}%
{driver \& OS}

\bugdesc%
{21}%
{On an arm and hand system, with hand and arm running on separate computers linked via an Ethernet bus, timestamped data got desynchronized.
This was noticed on the internal proprioception when fingers moved on the display and the arm did not although both moved in physical world.}%
{NTP not setup correctly.
University had a specific NTP setting requirement that was not set on some computers.
Could actually never synchronize.}%
{Looking at timestamps in the messages over \texttt{rosbag} or \texttt{rostopic} tools.
Analysing system clock drift with command line tools.}%
{working memory and CPU would be used more due to more interpolation\fshyp{}extrapolation computation between unsynced data streams.}%
{configuration}

\bugdesc%
{32}%
{PCL segfaulted on non-Debian\fshyp{}Ubuntu machines when trying to compute a convex hull.}%
{The code was written to support Debian's \texttt{libqhull}, ignoring the fact that Debian decided to deviate from the upstream module in one compile flag that changed a symbol in the library from \texttt{struct} to \texttt{struct*}.
That way all non-Debian ports of \texttt{libqhull} failed to work with PCL, and instead segfaulted while trying to access the pointer.}%
{\begin{itemize}
    \item minimal example
    \item \texttt{printf} within the PCL code
    \item \texttt{printf} within an overlayed version of \texttt{libqhull}
    \item \texttt{gdb}
    \item Debian package build description for \texttt{libqhull}
    \item upstream \texttt{libqhull} package
    \item 12 hours of continuous debugging.
\end{itemize}}%
{Well, segfault, the entire module stopped working. So basically everything was affected to some degree..}%
{driver \& OS; basic programming issue}

\bugdesc%
{46}%
{The robot kept asking someones name.}%
{Background noise in the microphone}%
{The bug was obvious: no limit on the amount of questions asked.
Simply drawing\fshyp{}viewing the state machine made this very obvious.}%
{}%
{coordination; environment-related}

\bugdesc%
{60}%
{signal processing in component chain gave different results after several months}%
{unknown}%
{}%
{}%
{not\fshyp{}accidentally solved}

\bugdesc%
{69}%
{segfault}%
{timing and location of allocated memory}%
{memory dumps\dots many many memory dumps}%
{it did not affect resources constantly, but system stability in general; maybe CPU and memory}%
{basic programming issue}

\bugdesc%
{76}%
{While operating, a robot system normally capable of autonomous obstacle avoidance would unexpectedly drop communication with its wireless base station and drive erratically with high probability of collision.}%
{The main process was started in a Linux terminal and launched a thread that passed wheel velocity information from the main process to the robot controller. When the terminal was closed or otherwise lost, the main process was terminated but the thread continued to run, supplying old velocities to the robot controller.}%
{top, debugger, thought experiments}%
{None}%
{coordination}

\bugdesc%
{83}%
{Random segfaults throughout system execution.}%
{Bad memory allocation: \texttt{malloc} for \texttt{sizeof(type)} rather than \texttt{sizeof(type*)}.}%
{Backtrace with \texttt{gdb}, profiling with \texttt{valgrind}, eventual serendipity to realize the missing \texttt{*} in the code.}%
{Memory}%
{basic programming issue}

\bugdesc%
{133}%
{memory mismatch, random crashes}%
{different components using different \texttt{boost} versions}%
{\texttt{debugger}, \texttt{printf}.
Finally solved after hint from colleague}%
{}%
{basic programming issue}

\bugdesc%
{149}%
{Erratic behaviour of logic}%
{Error in mathematical modeling}%
{Unit tests}%
{None}%
{}

\bugdesc%
{150}%
{An algorithm was implemented in both C++ and MATLAB exactly the same way.
However, only the MATLAB implementation was working correctly.}%
{Difference in storing the float point variables in MATLAB and C++.
MATLAB rounded the numbers, however, C++ cut them.}%
{Step by step tracing and debugging, and watching variables.
Then, comparing with each other.}%
{Working memory}%
{basic programming issue}

\bugdesc%
{153}%
{Control Program crash after a consistent length of time}%
{Presumably memory leak.
Never knew for sure.}%
{Not sure}%
{Not sure}%
{basic programming issue; performance bug}

\bugdesc%
{156}%
{Visualization window crashing \SI{100}{\percent} of the time I open it.
Running the program inside of \texttt{gdb} resulted in the program successfully running \SI{100}{\percent} of the time.}%
{??? Likely something internal to closed-source graphics drivers interacting with OpenGL\fshyp{}OGRE}%
{Was able to eventually generate a backtrace that pointed to graphics drivers.}%
{CPU\fshyp{}Memory\fshyp{}GPU were all affected because I had to run the program inside of \texttt{gdb}}%
{driver \& OS}

\bugdesc%
{162}%
{bad localization of a mobile robot in outdoor campus environment.
Jump of the estimation}%
{Bad wheel odometry reading.}%
{Analyze log file}%
{None. Loss of performance due to incorrect position tracking}%
{}

\endgroup

\sloppy
\printbibliography[nottype=software]{}
\printbibliography[type=software,title={Software packages}]{}
\fussy

\end{document}